\def\year{2021}\relax
\documentclass[letterpaper]{article} % DO NOT CHANGE THIS
\usepackage{aaai21}  % DO NOT CHANGE THIS
\usepackage{times}  % DO NOT CHANGE THIS
\usepackage{helvet} % DO NOT CHANGE THIS
\usepackage{courier}  % DO NOT CHANGE THIS
\usepackage[hyphens]{url}  % DO NOT CHANGE THIS
\usepackage{graphicx} % DO NOT CHANGE THIS
\usepackage{natbib}  % DO NOT CHANGE THIS AND DO NOT ADD ANY OPTIONS TO IT
\usepackage{defs}
\usepackage{amsmath}
\usepackage{booktabs}
\usepackage{pifont}%
\newcommand{\cmark}{\ding{51}}%
\newcommand{\xmark}{\ding{55}}%
\usepackage{caption} % DO NOT CHANGE THIS AND DO NOT ADD ANY OPTIONS TO IT
\usepackage{xcolor}
\usepackage{latexsym}
\usepackage{multirow}
\usepackage[switch]{lineno}
\usepackage{appendix}
\definecolor{mgreen}{rgb}{0.0, 0.5, 0.0}
\definecolor{mblue}{rgb}{0.19, 0.55, 0.91}
\definecolor{cb_orange}{rgb}{1.0,0.51,0.0}
\definecolor{cb_blue}{rgb}{0.22,0.49,0.72}
\definecolor{cb_green}{rgb}{0.3,0.67,0.29}

\usepackage{enumitem}

\usepackage{hyperref}

\makeatletter
\newcommand{\printfnsymbol}[1]{%
  \textsuperscript{\@fnsymbol{#1}}%
}
\makeatother
\title{Generate Your Counterfactuals: Towards Controlled\\Counterfactual Generation for Text}
\author{
    Nishtha Madaan,
    Inkit Padhi,
    Naveen Panwar,
    Diptikalyan Saha
    \\
}
\affiliations{
    \textsuperscript{\rm 1}IBM Research AI\\
    \{nishthamadaan, naveen.panwar, diptsaha\}@in.ibm.com, inkpad@ibm.com

}
\begin{document}
%\linenumbers  

\maketitle

\begin{abstract}
Machine Learning has seen tremendous growth recently, which has led to a larger adoption of ML systems for educational assessments, credit risk, healthcare, employment, criminal justice, to name a few. Trustworthiness of ML and NLP systems is a crucial aspect and requires guarantee that the decisions they make are fair and robust. Aligned with this, we propose a framework \textit{GYC}, to generate a set of counterfactual text samples, which are crucial for testing these ML systems. Our main contributions include a) We introduce GYC, a framework to generate counterfactual samples such that the generation is plausible, diverse, goal-oriented and effective, b) We generate counterfactual samples, that can direct the generation towards a corresponding \texttt{condition} such as named-entity tag, semantic role label, or sentiment. Our experimental results on various domains, show that GYC generates counterfactual text samples exhibiting the above four properties. %The generated counterfactuals can then be fed complementary to the existing data augmentation for improving the  debiasing  algorithms performance  as  compared  to  existing  counterfactual  generated by token substitution. 
GYC generates counterfactuals that can act as test cases to evaluate a model and any text debiasing algorithm. 

%\footnotetext{Code and implementation details will be available at: https://github.com/annon-author9/GYC }

\end{abstract}
%\nm{More cleaner examples}
\begin{table*}
\centering
\scriptsize
\begin{tabular}{@{}|p{0.19\textwidth}|p{0.22\textwidth}|p{0.23\textwidth}|p{0.22\textwidth}|@{}}
% \toprule
\hline
\textbf{Input Sentence}  & \begin{tabular}[c]{@{}p{0.25\textwidth}@{}} \textbf{Token-based Substitution}\\ \cite{ribeiro2020beyond}, \cite{devlin2018bert}\\\end{tabular}    & \begin{tabular}[c]{@{}p{0.25\textwidth}@{}} \textbf{Adversarial Attack}\\  \cite{michel2019evaluation}\\\end{tabular}  & \begin{tabular}[c]{@{}p{0.25\textwidth}@{}} \textbf{GYC} \\(Ours) \\\end{tabular}                                                                              \\ \hline
I am very \textit{disappointed} with the service & \begin{tabular}[c]{@{}p{0.25\textwidth}@{}}I am very \textbf{pleased} with the service.\\ I am very \textbf{happy} with the service.\\  I am very \textbf{impressed} with the service.\end{tabular} & \begin{tabular}[c]{@{}p{0.20\textwidth}@{}}I am very \textbf{witty} with the service.\end{tabular}  & \begin{tabular}[c]{@{}p{0.25\textwidth}@{}} 
I am very \textbf{pleased to get a good} service. \\ 
I am very \textbf{happy} with \textbf{this} service. \\
I am very \textbf{pleased} with the service.
\end{tabular} \\
\hline 
%   What is the best time to visit Singapore? & \begin{tabular}[c]{@{}p{0.25\textwidth}@{}}What is the \textbf{right} time to visit Singapore?\\\\ What is the \textbf{perfect} time to visit Singapore?\\ \\ What is the \textbf{appropriate} time to visit Singapore?\end{tabular} & \begin{tabular}[c]{@{}p{0.25\textwidth}@{}}\textbf{overcooked} is the best time to visit Singapore?\end{tabular}  & \begin{tabular}[c]{@{}p{0.25\textwidth}@{}} What is not is the same as Singapore?\\\\ What is the most difficult time in Singapore?\\\\
% What is a neighboring problem of Singapore?  \end{tabular}                                                                                                                                \\ \midrule
% I enjoyed the food here. & \begin{tabular}[c]{@{}p{0.25\textwidth}@{}}I \textbf{like} the food here.  \\\\ I \textbf{love} the food here.\\ \\ I \textbf{want} the food here.\end{tabular} & \begin{tabular}[c]{@{}p{0.25\textwidth}@{}}\textbf{milquetoast} enjoyed the food here.\end{tabular}  & \begin{tabular}[c]{@{}p{0.25\textwidth}@{}} I hate the food here.\\\\ I did not like how the food. \\\\  
1st time burnt \textit{pizza} was \textit{horrible}. &
\begin{tabular}[c]{@{}p{0.25\textwidth}@{}}
1st time burnt \textbf{house} was \textbf{destroyed}. \\ 
1st time burnt \textbf{place} was \textbf{burned}.\\  
1st time burnt \textbf{house} was \textbf{burnt}.
\end{tabular} 

& \begin{tabular}[c]{@{}p{0.25\textwidth}@{}}\textbf{personable} time burnt pizza was horrible. \end{tabular}  &
\begin{tabular}[c]{@{}p{0.25\textwidth}@{}} 
1st time burnt pizza \textbf{is delicious}.\\
1st time burnt \textbf{coffee} was \textbf{delicious}. \\  
1st time burned pizza was \textbf{delicious}.
\end{tabular}
% \\ \bottomrule
\\ \hline
\end{tabular}
\caption{Comparison of generated counterfactual text from existing and proposed model on text from YELP dataset.}
\label{tab:example_samples_related}
\end{table*}

\noindent 

\section{Introduction}

% Grand challenge

%\ip{need for test suite}
Owing to the advancement of deep learning in recent years, AI systems are now deployed for varied tasks and domains like educational assessments, credit risk, healthcare, employment, criminal justice. With the deployment of NLP systems in these domains, the need to ascertain trustworthiness and reliability becomes crucial to ensure the decisions that they make are fair and robust. To this end, counterfactual text \cite{pearl2000models} provides an alternate dataset which acts as test-cases to evaluate the properties such as fairness and robustness of these systems. These texts along with its correct labels may also be used to augment the training set to mitigate such properties  \cite{garg2019counterfactual}. %Counterfactual examples play an important role in evaluating machine learning model as these counterfactual provide valuable insights about the decision-making power of ML models.
%Thus generating plausible, diverse and goal-oriented counterfactuals becomes an important task to ensure realistic test cases with high coverage of the input space defined by the goal.  

The four properties that are important for counterfactual text are plausibility, diversity, goal-orientedness, and effectiveness. Plausibility ensures that the generated test cases are realistic and can occur in the input and thus can be further used for retraining. Goal-orientedness can ensure that the generated counterfactual text samples are deviating from the original sample on a particular aspect like NER, SRL, synonyms or fairness attributes as controlled by the user. The diversity will ensure high coverage of the input space defined by the goal. In this paper, we aim to generate such counterfactual text samples which are also effective in finding test-failures (i.e. label flips for NLP classifiers).

% \nm{Recent advancements in test-cases world towards fairness, robustness/ adverserial} 

Recent years have seen a tremendous increase in the work on fairness testing \cite{mametamorphic,holstein2019improving} which are capable of generating a large number of test-cases that capture the model's ability to misclassify or remove unwanted bias around specific protected attributes \cite{huang2019reducing}, \cite{garg2019counterfactual}. This is not only limited to fairness but the community has seen great interest in building robust models susceptible to adversarial changes \cite{goodfellow2014explaining, michel2019evaluation, ebrahimi2017hotflip, zhao2017generating}. These models are capable of generating sentences that can change the label of the classifier. These works are very limited to a specific domain. While the change of label is likely, the generations are not plausible and are ungrammatical \cite{li2016understanding} or require manual human intervention in many cases \cite{jia2017adversarial}.

%\nm{Checklist, What is our goal?} 
A recent work by \citealp{ribeiro2020beyond} employs a tool Checklist which is one of the attempts to come up with generalized test-cases. For generation, Checklist uses a set of pre-defined templates, lexicons, general-purpose perturbations, and context-aware suggestions. A major limitation with these template-based approaches or rule-based approaches is that these can not generate meaningful diverse counterfactual samples. Our goal is to be able to generate these counterfactual samples for a given text in a more generalized manner with high generation quality and diversity. 

% How transformers are capable of generating plausible sentences and scope out our work
Generative models like GPT-2 \cite{radford2019language} are good at generating plausible and diverse text. We can leverage these capabilities to generate counterfactual text \cite{pearl2000models}. However, directly adopting Transformers architecture \cite{NIPS2017_7181} to generate perturbations for a goal-oriented task is difficult without adding additional attributes and fine-tuning with attribute-specific data \cite{keskar2019ctrl}.
Similar to \cite{dathathri2019plug}, we perform controlled text generation to achieve the above objectives. There are two key technical challenges in adopting this approach. First, transformers \cite{NIPS2017_7181} are good at generating feasible and plausible but inferencing on GPT-2 is hard. Second, prior approaches to control GPT-2 text generation do not support a custom model to direct the generation.
%Introducing our framework
% Rough solution sketch
In this paper, we address the drawbacks of the above approaches by keeping the balance of the four properties of counterfactual text.  This framework can be used to plug in any custom model to direct the generation towards counterfactual. In our framework, we first reconstruct the input text and then introducing losses that work with differential and non-differentiable models to enforce a corresponding \texttt{condition}. This condition can be a sentiment, NER, or a different user-provided class label. The differentiable loss is computed over logits of the input sentence, while non-differentiable loss relies on generating reward computed over text. Further, to ensure that the generated counterfactual text is diverse, we add entropy over logits of input.

% Problems with current approaches and how are we addressing these. 
 Few examples of given input sentences in Table \ref{tab:example_samples_related} show that the existing methods generate hard test cases with either single word replacements or by adding an existing template to generate multiple test-cases which do not possess the required generation quality and are neither diverse in nature. Recent adversarial approaches like \cite{michel2019evaluation} generate adversarial changes on input text to change the label of output sentence, regardless of whether the output sentence makes sense. These adversarial generations may flip the label but these generations are not likely to be ever seen in user input. Therefore these test-cases become ineffective. On the other hand, it would be more useful to generate counterfactual text samples that are likely to be seen in user inputs.
% Technical challenges: Specifically what is the problem
%  We argue that test cases for a given sentence $\bx$ can be generated by perturbing the hidden states $\bH_t$, so that a corresponding  $\by^k$ can be generated. First, it is non-trivial to sample different counterfactuals to obtain test-cases. Directly adopting Transformers architecture \cite{NIPS2017_7181} to generate perturbations for a specific model is difficult without adding additional attributes and fine-tuning with attribute-specific data \cite{keskar2019ctrl}, \cite{ziegler2019fine}, Second, we can hardly obtain meaningful counterfactuals by directly substituting template-based tokens or rule-base substitutions \cite{ribeiro2020beyond}. 

% Contributions 
In particular, our key contributions are listed as follows:
\noindent
\begin{enumerate}[label=\textbf{(\alph*)}, wide = 0pt, labelindent=0pt]
  \item We introduce GYC, a framework to generate counterfactual samples such that the generation is plausible, diverse, goal-oriented, and effective. This is in contrast to the previous work which produces adversarial text but is ungrammatical or non-plausible. We show a comparison in examples 1 and 2 in Table \ref{tab:example_samples_related}.
    \item We generate counterfactual samples, that can direct the generation towards a corresponding \texttt{condition} such as NER, SRL, or sentiment. We quantify our results (Tab. \ref{tab:joint_comparison_main} and  \ref{tab:joint_comparison_aux}) using automated evaluation metrics (by training \texttt{condition} models) to check Label-flip and diversity. Through human judgment we measure plausibility, grammar check, and validity of generation.
    \item GYC allows plugging-in any custom \texttt{condition} model to guide the counterfactual generation according to user-specific \texttt{condition}. We show that our framework is able to conditionally generate counterfactual text samples with high plausibility, label-flip score, and diversity. 
    \item GYC enables the use of pre-trained GPT-2 decoder and does not require any re-training or fine-tuning of the model.  
\end{enumerate}
% \noindent \textbf{Outline.}
% The rest of the paper is structured as follows: In Section 2 we talk about the background and building blocks for our main method. Further in Section 3, we describe the method for counterfactual generation. 
% In Section 4, we briefly discuss applications of these generated counterfactuals. Further in Section 5, we talk about related work in text generation, counterfactual generation and adversarial attacks. In Section 6, we show experiment results with baselines. In Section 7 we conclude the paper.

\section{Background}
% Starts at 1.25
% \subsection{Notations}
% \begin{align}
%     t && \text{token-step} \\
%     T && \text{Text length} \\
%     k && \text{sample index} \\
%     K && \text{Number of samples} \\
%     \bx && \text{Input Text}\\
%     \by^k && \text{Counterfactual Text Sample}\\
%     \by_t && \text{Token in Counterfactual Text} \\
%     \boldsymbol{o}_t && \text{Logits}\\
%     \bH_t && \text{History Matrix} \\
%     \Delta\bH_t && \text{Perturbation}
% \end{align}
\subsection{Language Modeling for Text Generation}
Language Models have played a significant role through introduction of word2vec \cite{w2vec}, contextual word vectors \cite{devlin2018bert} and deep contextualized LM models \cite{radford2019language, howard2018universal}. 

Recent advances by \citealp{NIPS2017_7181} introduce large transformer-based architectures which have enabled training on large amounts of data \cite{radford2019language, devlin2018bert}

In this modeling, there is a history matrix that consists of key-value pairs as $\bH_t=[(K_{t}^{(1)},V_{t}^{(1)}), \ldots, (K_{t}^{(l)},V_{t}^{(l)})]$, where $(K_{t}^{(i)},V_{t}^{(i)})$ corresponds to the key-value pairs of the $i-$th layer of a transformer-based architecture at time-step $t$. Then, we can efficiently generate a token $\by_{t}$ conditioned on the past text $\by_{<t}$. This dependence on the past text is captured in the history matrix $\bH_t$. Using this model a sample of text can be generated as follows.
\begin{align}
    \bo_{t+1}, \bH_{t+1} &= \operatorname{LM}(\by_t, \bH_t) \\
    \by_{t+1} &\sim \operatorname{Categorical}(\bo_{t+1}),
    \label{eq:gpt2lm}
\end{align}
Here, $\bo_{t+1}$ are the logits to sample the token $\by_{t+1}$ from a Categorical distribution.

\subsection{Controlled Text Generation}
The aim of controlled text generation is to generate samples of text conditioned on some given \texttt{condition}.
\begin{align}
    p(\by | \texttt{condition})
\end{align}
This condition can be a class label such as the sentiment (\textit{negative} or \textit{positive}), topic (\textit{sci-fi} or \textit{politics}) or tense (\textit{past}, \textit{present} or \textit{future}) \citep{hu2017toward, dathathri2019plug}. The condition can also be structured data such as a Wikipedia infobox or even a knowledge graph \citep{ye2020variational}.

\citet{dathathri2019plug} propose a new framework called PPLM for controlled text generation. PPLM uses a pre-trained GPT-2 to generate text as stated in eq. \ref{eq:gpt2lm}. However, to control the text, PPLM perturbs the history matrices $\bH_t$ by adding a learnable perturbation $\Delta \bH_t$ to get a perturbed history matrix $\tilde{\bH_t}$. 
\begin{align}
    \tilde{\bH}_t = \bH_t + \Delta \bH_t
\end{align}
This is plugged into eq.~\ref{eq:gpt2lm} to generate the controlled samples.
\begin{align}
    \bo_{t+1}, \bH_{t+1} &= \operatorname{LM}(\by_t, \tilde{\bH}_t) \\
    \by_{t+1} &\sim \operatorname{Categorical}(\bo_{t+1}),
    \label{eq:pplm}
\end{align}
PPLM approximates the process of sampling from the required distribution for $p( \by | \texttt{condition})$. For this, PPLM trains the learnable parameters $\Delta \bH_t$ so that perturbed samples approximate the required distribution. This is done by maximizing the log-probability of $p( \by | \texttt{condition})$.  To evaluate this distribution, Bayes rule is used as follows:
\begin{align}
    p(\by | \texttt{condition}) \propto p(\by) p( \texttt{condition} | \by)
\end{align}
For instance, to enforce the condition so that generated text samples belong to a class $c$ with respect to some given classifier $p(c|\by)$, PPLM maximizes the following objective.
\begin{align}
    \log p(c | \by) - \sum_t\KL\left[p(\by_t| \by_{t-1}, \tilde{\bH}_{t-1}) || p(\by_t| \by_{<t})\right]
\end{align}

Here, the first term enforces the class label. The second term is a KL term which enforces that the text is plausible and is in line with unperturbed GPT-2 distribution.
% To learn the perturbations $\Delta \bH_t$, PPLM minimizes an objective  
% PPLM uses a pre-trained GPT-2 to generate attribute-specific text samples. 
% Let $\Delta H_t$ be the update to $H_t$, such that the generation with $(H_t + \Delta H_t)$ makes the generated text more likely to possess the desired attribute. For a pre-specified number of iterations $L$, PPLM shifts the history $H_t$ in the direction of the sum of the gradients $\nabla_{H_t} \log p(a|x)$ and $\nabla_{H_t} \log p(x)$.  Then, the PPLM creates a perturbed text by sampling token $\tilde{x}_i$ from equation (\ref{eq:pplm}).
% \begin{align}
%     \tilde{\bo}_{t+1}, \bH_{t+1} &= \operatorname{LM}(\bx_t, \tilde{H}_t) \\
%     \tilde{x}_{t+1} &\sim \operatorname{softmax}(W \tilde{o}_{t+1}),
%     \label{eq:pplm}
% \end{align}
% where $\tilde{H}_t = H_t + \Delta H_t$. After $L$ such perturbations the result is the desired text.

\subsection{Counterfactual Text} \label{ssec:num1}
  A counterfactual example is defined as synthetically generated text which is treated differently by a \texttt{condition} model. Any text is a description of a scenario or setting. For instance, given a text $\bx$, \textit{My friend lives in London}, a counterfactual text $\by$  becomes - \textit{My friend lives in California}, if the friend counterfactually lived in California. From the perspective of a given classifier model, a counterfactual example defines how the original class label flips if the friend lived in California instead of London. Such generated counterfactuals can serve as test-cases to test the robustness and fairness of different classification models. Some examples can be seen in Table \ref{tab:example_perts}. An effort to make the classifier fairer could utilize such counterfactual data for data augmentation task \cite{garg2019counterfactual}.  

\begin{table}[t]
\small
\begin{tabular}{@{}|p{0.45\textwidth}|p{0.45\textwidth}|@{}}

% \toprule

\hline
\\
\textbf{Model} : NER Location Tagger\\
\textcolor{mblue}{Source}: N/A, \textcolor{mgreen}{Target}: Perturb Location-Tag \\ \\
\textcolor{mblue}{Input Sentence}: \textit{My friend lives in beautiful London.} \\ 
\textcolor{mgreen}{Counterfactual Text Samples} :\\
{[}1{]} My friend lives in majestic downtown Chicago. \\ {[}2{]}  My friend lives in gorgeous London. \\ {[}3{]} My friend lives in the city of New Orleans.\\ \\

% \midrule
\hline
\\
\textbf{Model} : Sentiment Classifier\\
\textcolor{mblue}{Source}: Negative Class Label, \textcolor{mgreen}{ Target}: Positive Class Label \\ \\
%\\ \midrule
\textcolor{mblue}{Input Sentence}: \textit{I am very disappointed with the service.} \\
\textcolor{mgreen}{Counterfactual Text Samples} :\\
{[}1{]} I am very pleased with the service.\\ {[}2{]} I am very happy with this service \\ {[}3{]} I am very pleased to get a good service.\\ \\
\hline
\\
\textbf{Model} : Topic Classifier\\
\textcolor{mblue}{Source Topic}: World, \textcolor{mgreen}{Target Topic}: Sci-Fi \\ \\
\textcolor{mblue}{Input Sentence}: \textit{The country is at war with terrorism.} \\ 
%\\ \midrule
\textcolor{mgreen}{Counterfactual Text Samples} :\\
{[}1{]} The country is at war with piracy at international waters.
\\ {[}2{]} The country is at war with its own beurocracy.  \\ 
{[}3{]} The country is at war with piracy offenses. \\ \\

% \bottomrule 
\hline
\end{tabular}
\caption{Illustration of samples generated by GYC. In the first example, \texttt{condition} enforced is to change the location tag. GYC selectively changes the city \textit{London} to \textit{Chicago} and \textit{New Orleans} while maintaining the sense of the remaining text. In the second example, the \texttt{condition} enforced is to change the sentiment from negative to positive with respect to a provided classifier. In the third example, the goal is to change the topic label of the text to \textit{sci-fi}. }
\label{tab:example_perts}

\end{table}

% In the next section, we describe our main approach to generate these counterfactual samples.
% Setup notations

% Basic GPT-2
% PPLM
% DICE

\section{GYC Method}
% Starts at page 2.5
In this section, we describe our method for generating $K$ counterfactual text samples $Y = \{ \by^k \}_{k=1}^K$ given a text $\bx$ and controlling \texttt{condition} that specify the scope of the counterfactual text. Formally, this modelling is described by: 

\begin{align}
    p(\by|\bx, \texttt{condition})
\end{align}

To implement this, we first obtain $\tilde{\bH}$ which depends on both the \texttt{condition} and the $\bx$ as opposed to just the condition. Hence, we first compute $\tilde{\bH}$ on the basis of $\bx$ such that the generated samples reconstruct the input text $\bx$. Next, the \texttt{condition} is enforced by further perturbing the $\tilde{\bH}$ to obtain $\hat{\bH}$.  
% In this work, we use Transformer \cite{NIPS2017_7181} to model the natural language. As shown in Figure \ref{fig:1}, we take the following steps to obtain counterfactual samples : 

% \begin{enumerate}
%     \item Given input text $x$, we infer the GPT-2 hidden state $h$.
%     \item For $h$ we generate $K$ perturbations $\{\Delta_k\}_K$ which gives $K$ counterfactual hidden states $\{H_k\}_K$.
%     \item We use GPT-2 to decode the perturbed hidden states to get $\{x_k\}_K$ counterfactual samples.
%  \end{enumerate}
\begin{figure}[ht]
    \centering
    \includegraphics[width=1.0\linewidth]{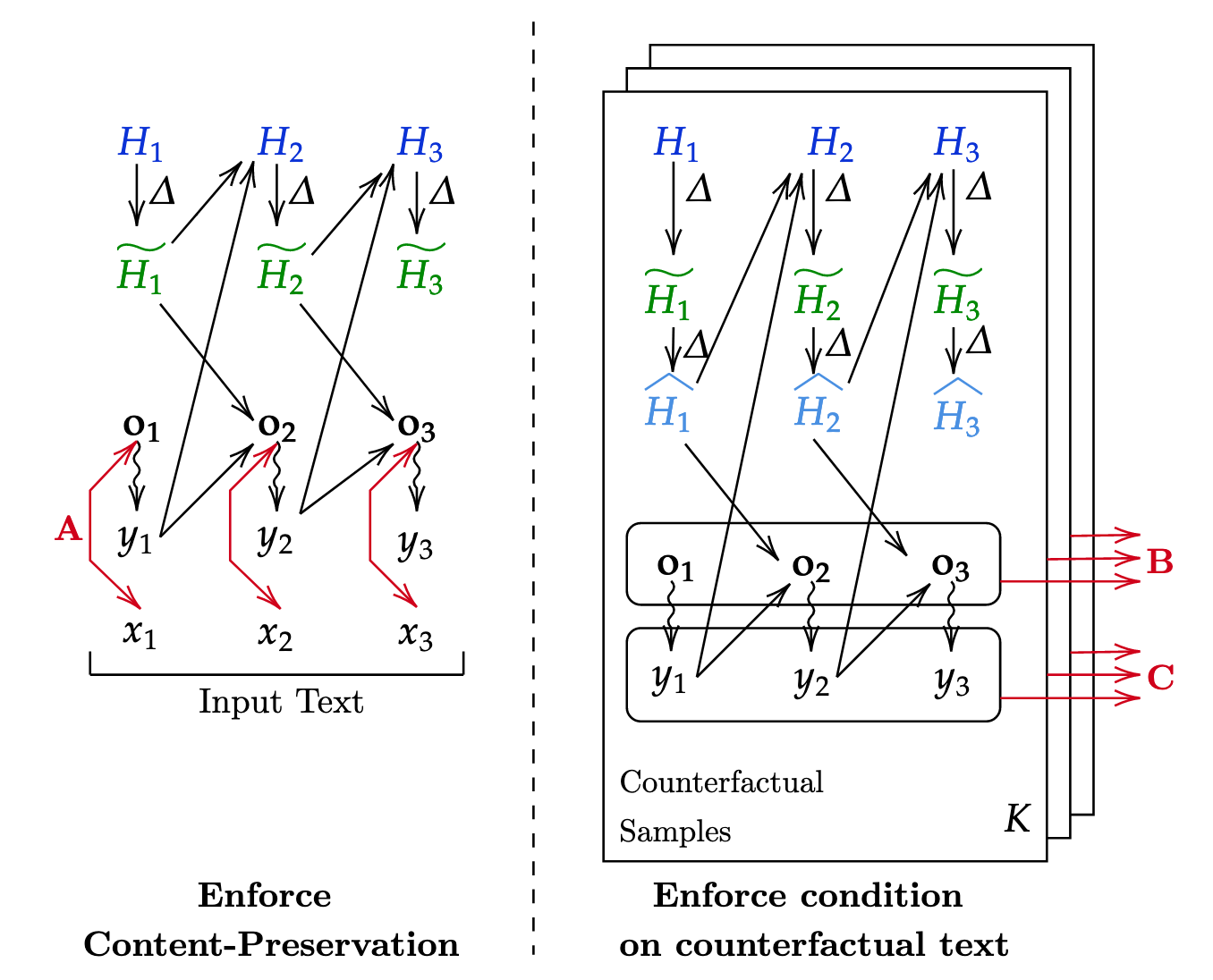}
    \caption{The architecture diagram showing interplay of different losses for generation. \textcolor{red}{\textbf{A}} represents proximity loss,\\ \textcolor{red}{\textbf{B}} represents differentiable score and entropy, \textcolor{red}{\textbf{C}} represents reward based score and entropy, $\Delta$ represents perturbation. }
    \label{fig:gyc}
\end{figure}
\subsection{Generative Process}
Formally, our aim is to draw $K$ counterfactual samples as $\by^k \sim p(\by|\bx, \texttt{conditions})$. This can be factorized auto-regressively over time as:
\begin{align}
    p(\by|\bx, \texttt{condition}) = \prod_{t=1}^T p(\by_t | \by_{<t}, \bx, \texttt{condition})
\end{align}
To implement this, we learn $\hat{\bH}$ on the basis of the given text $\bx$ and the given \texttt{condition} and use it to sample the counterfactual text samples as follows:
\begin{align}
    \bo_{t+1}, \bH_{t+1} &= \operatorname{LM}(\by_t, \hat{\bH}_t) 
    \label{eq:cge0}\\
    \by_{t+1} &\sim \operatorname{Categorical}(\bo_{t+1}),
    \label{eq:cge1}
\end{align}

\subsection{Learning Counterfactuals}
In this section, we describe our approach for learning the hidden state perturbations $\tilde{\bH}_t$ for reconstruction and subsequently $\hat{\bH}_t$ for enforcing the given conditions. Note that tilde on $\tilde{\bH}_t$ represents perturbation for reconstruction and hat on $\hat{\bH}_t$ represents the perturbation for condition-oriented generation.  We describe the process through the following loss functions used in our generation process also shown in Fig \ref{fig:gyc}. 

\subsubsection{Reconstruction via Proximity Loss.} 
A key problem in generative models is to perform inference over the latent states given the observation. GPT-2 suffers from the same problem. In our setup, the inference step is crucial because we require the latent state that generates the given input text before we can perturb it to generate counterfactual text.

To do this, we first aim to learn perturbations $\Delta \bH_t$ such that the generated samples reconstruct the given text $\bx$. This is done by learning $\Delta \bH_t$ to maximize the log probability of the given text.
\begin{align}
   \mathcal{L}_{p} &= \log p(\by = \bx|\bx, \texttt{condition})\\
   &= \sum_t \log p(\bx_t | \bx_{<t}, \tilde{\bH}_{t-1})
   % = \prod_{t=1}^T p(\bx_t | \bx_{<t}, \bx, \texttt{condition})
\end{align}
% \begin{align}
%     \mathcal{L}_{p} \ =\ \sum \log p\left( \bx_{t} |\ \bx{< t}\right) 
% \end{align}

In a sense, this objective enforces proximity of the generated text to the original text and also enforces content preservation. Hence, we call it the proximity loss. For implementing this, we found a curriculum-based approach to perform better where we reconstruct one token at a time from left to right. For more details on this implementation, see Appendix A in supplementary material. %\ref{ax:proximity}.

\subsubsection{Enforcing \texttt{condition}}
To enforce \texttt{condition}, we assume we have access to a score function that takes the text and returns the degree to which the condition holds in the text. This score function may be differentiable with respect to the logits of the generated text, such as when the score is the probability of a class label returned by a model. In other cases, the score function may be akin to a reward and may not be differentiable with respect to the text. In either case, we aim to generate text that maximizes the score. To handle the differentiable and non-differentiable cases, we use the following losses.\\

\noindent
\textbf{Differentiable Score.}  Differentiable Score assumes that the input to the score function \texttt{score} are the logits $\bo_{1:T}$ of the tokens of the generated text. This can be translated into a corresponding differentiable loss as follows: 
\begin{align}
    \mathcal{L}_{d} = \texttt{score}(\bo_{1:T})
\end{align}
where $\bo_{1:T}$ are obtained from equation~\eqref{eq:cge0}.\\
For our experiments, we train a sentiment classifier on the datasets- Yelp \cite{shen2017style}, DBpedia \cite{dbpedia} and AgNews \cite{zhang2015character} which is used to compute the differentiable loss.\\

\noindent\textbf{REINFORCE.} To enforce a condition which is not differentiable with respect to the logits $\bo_{1:T}$ of the generated text, we resort to the REINFORCE technique. This type of scores are important because a variety of models take text as input and not the logits. For instance, a classifier may take text as input to perform tagging (such as NER or SRL) or to perform classification. We assume that the score function \texttt{score} is a reward that is obtained for the generated text. We draw $K$ samples of text $\{\by_{1:T}^k\}_{k=1}^K$ based on the currently learned $\hat{\bH}_t$. Then we compute a REINFORCE objective as follows.
\begin{align}
    \mathcal{L}_{r} \ =\sum ^{K}_{k\ =1} r^{k}\sum ^{T}_{t\ =1}\log \ p\left( \by^{k}_{t} |\by^{k}_{t-1}, \hat{\bH}_t\right)
\end{align}
where $r^k$ represents reward for each $k$ counterfactual text.
For our experiments, we use BERT-NER based model and SRL model to generate reward on NER and SRL in generated counterfactual text samples.\\

\noindent
\textbf{Diversity.}
To make sure that the distribution over the generated text does not collapse to the input text $\bx$ and to ensure diversity in the generated counterfactual text samples, we introduce a diversity loss. To implement this, we compute entropy of the logits of the $K$ generated text samples.
% \begin{align}
% \mathcal{L}_{H} \ =-\sum ^{K}_{k\ =1}\sum ^{T}_{t\ =1} \ \sum _{v\ \in \ V} p\left( v|x^{k}_{< t}\right)\log \ p\left( v|x^{k}_{< t}\right)\\
% \end{align}
\begin{align}
      \mathcal{L}_{H}  = \sum^{K}_{k=1}\sum^{T}_{t=1}  {H}\left(\by^{k}_{t} |\by^{k}_{<t}\right) \text{where } {H} \text{ is entropy}
\end{align}

\noindent
\textbf{Combined Objective.}
The combined objective which we maximize can be stated as following:
\begin{align}
    \mathcal{L} =\lambda _{r} \mathcal{L}_{r} + \lambda _{H} \mathcal{L}_{H} +\lambda _{p} \mathcal{L}_{p} 
\end{align}
where $\displaystyle \lambda _{r}$, $\displaystyle \lambda _{H}$ and $\displaystyle \lambda _{p}$ are hyper-parameters that can be tuned. To facilitate training we perform annealing on these hyper-parameters. In early training, we keep $\displaystyle \lambda _{r}$, $\displaystyle \lambda _{H}$ = 0. After reconstruction completes, $\displaystyle \lambda _{p}$ is lowered and  $\displaystyle \lambda _{r}$ and $\displaystyle \lambda _{H}$ are set to non-zero. See more details in Appendix A in supplementary document.

\section{Applications of Counterfactuals}
\subsection{Counterfactuals as Test-Cases}

Test-Cases are important in testing NLP models to perform a behavioral check on a model. These test-cases also complement the traditional test cases designed in the software engineering world and also seems to be very relevant with the increased adoption of NLP algorithms.  
GYC can be easily adapted to test any classification model with even a non-differentiable score function. One can plug any score function, and generate test-cases around a specific condition. In experiments, we take a score function as a Sentiment prediction function.  
GYC reports the highest scores in quantitative and qualitative results on different metrics across various datasets.  
% GYC reports average of 88.1\% self-bert score on Yelp dataset \cite{shen2017style}, which is a measure of diversity of generated test-cases. Also, it exhibits high label-flip score of 70.29\% which is a measure of the generated counterfactuals belonging to a different class than the source class which is very promising. 
The detailed experiments on sentiment and the performance on different metrics around generation and diversity are described in Table \ref{tab:joint_comparison_main} and \ref{tab:joint_comparison_aux} and Fig.~\ref{fig:human_eval}.

\subsection{Counterfactuals for Debiasing}

Applications of counterfactual text in debiasing have been studied well in literature by \cite{garg2019counterfactual}, \cite{madaan2018analyze} highlight multiple text debiasing algorithms which can be used to debias classification systems. The key to Data Augmentation and Counterfactual Logit Pairing Algorithms is that they require counterfactual text samples of the training data. Similarly, the counterfactual text is used to debias Language Model for Sentiment \cite{huang2019reducing}. These samples should satisfy a \texttt{condition} in this case a protected attribute should be present. While these techniques are claimed to be highly successful, in real-world getting enough data for a corresponding protected attribute is prohibitive. The performance of these algorithms can be boosted if the counterfactual text samples that are fed possess high generation quality. GYC can be used to generate data for such data augmentation tasks which can thereby help debias a model.

%  For this purpose, attribute specific counterfactual text can be generated which can thereby be fed to improve trust in any model. 
% We train a single layer classifier on the DBpedia data \cite{dbpedia} and show that with a similar hyper-parameter setting as used in GYC main model, counterfactuals generated using GYTC the flip rate after debiasing has been reduced to X on Y while with using token-substitution the flip rate was as high as Z on Y. 

%\subsection{Creative AI}

\section{Related Work}
% Themis (label-flip score), Equitus (label-flip score), Ours (diversity, label flip score), ICSE 2020 (SMU, white box models, label-flip score)
\subsection{Test-Case Generation}
Recent works on software testing have been addressing the problem of increasing fairness in different applications. In structured data, Themis \cite{udeshi2018automated}, Equitus \cite{galhotra2017fairness} and \cite{aggarwal2019black, john2020verifying} use random testing to generate test-cases. 
\cite{wachter2017counterfactual} defined the generation of counterfactual explanations that can be used to evaluate a model. Further \cite{mothilal2020explaining} address the feasibility and diversity of these counterfactual explanations. 
In text data, recently the work by \cite{ribeiro2018anchors} introduced a framework to generate test-cases relying on token-based substitution. This work is the closest work to our introduced framework.
    
% Heading 2 about some sub-field (Controlled text Generation)
    % Why our works is not conflicting with this
\subsection{Controlled Text Generation}
 
Due to the discrete and non-differential nature of text, controlled text generation task is non-trivial and most of the current methods indulge in fine-tuning using reinforcement \cite{yu2017seqgan}, soft argmax \cite{zhang2017adversarial} or Gumbel-Softmax techniques \cite{jang2016categorical}. Moreover, due to advancements in pre-trained text representations, recent work tries to achieve conditional generation without requiring the need to re-training. \cite{engel2017latent} enforces behavioral latent constraints as part of post-hoc learning on pre-trained generative models, whereas \cite{dathathri2019plug} incorporates attribute discriminators to steer the text generations.

\subsection{Adversarial Attack}
Adversarial literature especially the work by \citet{ebrahimi2017hotflip} have shown promising performance in generating samples that tend to change the classifier's outcome. \cite{ebrahimi2017hotflip} showed that by using a gradient-based method and performing a minimal change in the sentence the outcome can be changed but the generated sentences might not preserve the content of the input sentence. \cite{michel2019evaluation, zhao2017generating} explored ways to preserve content while also change the classifier's outcome. For instance, \cite{michel2019evaluation}, leverages a method of imbuing gradient-based word substitution with constraints aimed at preserving the sentence's meaning. %Motivated by the idea of generating perturbed samples, our GYC framework not only preserves the content of the generated counterfactual but also generates diverse counterfactual text.

\begin{table}[h]
\scriptsize
\centering
\begin{tabular}{@{}l|cccc@{}}
\toprule
         & \multicolumn{1}{l}{Checklist} & \multicolumn{1}{l}{PPLM-BoW} & \multicolumn{1}{l}{Masked-LM} & \multicolumn{1}{l}{GYC} \\ \midrule
Dictionary-free     & \xmark                                   & \cmark                           & \cmark  & \cmark                     \\
Sentence-Level CF      & \xmark                                  & \xmark                            & \xmark & \cmark                                \\
Diversity & \xmark & \xmark & \xmark & \cmark\\
Enforce \texttt{condition} & \xmark & \cmark & \xmark & \cmark\\

\bottomrule
\end{tabular}
 \caption{Models for evaluation.}
\label{tab:baselines}
\end{table}
% There have been various works in domain of adversarial attack where the goal is to change the classifier's outcome by performing a minimal change in the sentence. \cite{ebrahimi2017hotflip} describes a method to change the characters in a sentence while achieving maximum change in gradient which lead to the different label for the sentence. Changing the sentence based on such a technique might lead to change the meaning of the sentence as well. To overcome this challenge \cite{michel2019evaluation} has proposed a method of imbuing gradient-based word substitution with constraints aimed at preserving the sentence's meaning. 

% Heading 3 about some sub-field (Adversarial attacks)
    % Why our works is not conflicting with this   

% Start at 4.25

% Heading 1 about some sub-field (Structured data)
    % Why our works is not conflicting with this
    
% Heading 2 about some sub-field (Controlled text Generation)
    % Why our works is not conflicting with this
    
% Heading 3 about some sub-field (Adversarial attacks)
    % Why our works is not conflicting with this   

\section{Experiments}
\begin{table}[ht]
\centering
\begin{tabular}{lrccc}
\toprule
\multicolumn{1}{l}{Metric} & Model & Dbpedia &  Agnews & Yelp\\
\midrule
\multirow{4}{*}{\begin{tabular}[c]{@{}l@{}}  Label-\\Flip \\ Score \\ ($\uparrow$ better) \end{tabular}} 
& Masked-LM & 30.00 & 37.80 & 10.87 \\ 
& PPLM-BoW & \textbf{75.29} & 42.19 & 35.55 \\
& Checklist & 24.03 & 27.27 & 4.00 \\
& GYC & 72.09 & \textbf{42.76} &  \textbf{70.29} \\
\midrule
\multirow{4}{*}{\begin{tabular}[c]{@{}l@{}}  Diversity \\Score \\ ($\downarrow$ better) \end{tabular} } 
& Masked-LM & 0.98 & 0.97 & 0.96  \\
& PPLM-BoW & 0.88 & \textbf{0.88} & \textbf{0.87} \\
& Checklist & 0.98 & 0.97 & 0.97 \\
& GYC & \textbf{0.87} & 0.90 & 0.88 \\

\bottomrule
\end{tabular}
\caption{Evaluation of Label Flip Score (representing the counterfactual text samples generated with a different label) and Diversity Score (representing the diversity in generated counterfactual text) of baselines.}
\label{tab:joint_comparison_main}
\end{table}
% Small tables : Table naveen end 

\begin{figure}[ht]
    \centering
    \includegraphics[width=1.0\linewidth]{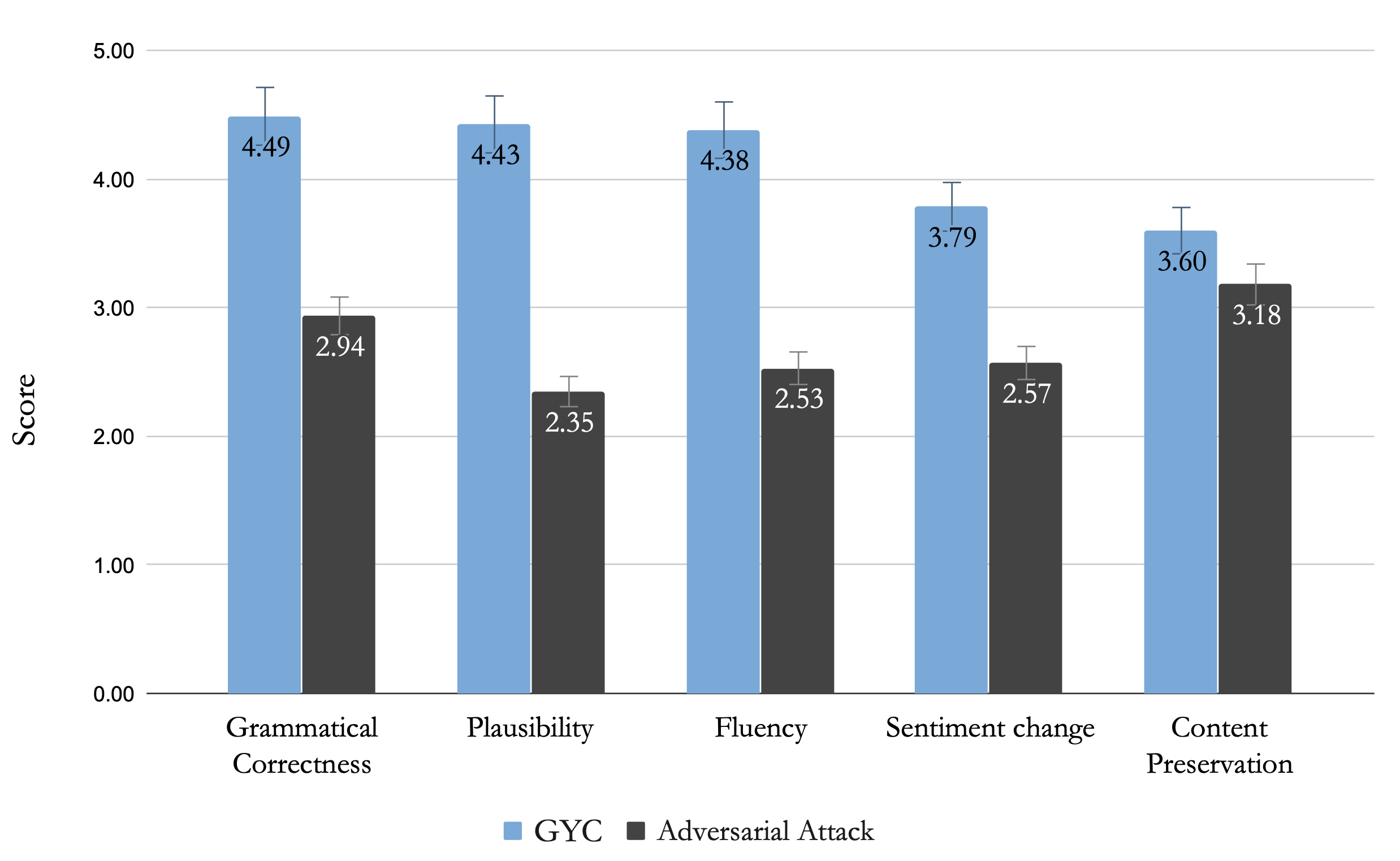}
    \caption{Comparison of human evaluation of generated text from Adversarial Attack and GYC.}
    \label{fig:human_eval}
\end{figure}

% Small tables : Table naveen-inspired table 
\begin{table}[ht]
\centering
\begin{tabular}{lrccc}
\toprule
\multicolumn{1}{l}{Metric} & Model & Dbpedia &  Agnews & Yelp\\
\midrule
\multirow{4}{*}{\begin{tabular}[c]{@{}l@{}}  Content \\ Preserv-\\ation \\ ($\uparrow$ better) \end{tabular}} 
& Masked-LM & 0.89 & 0.84 & 0.83 \\ 
& PPLM-BoW & 0.18 & 0.20 & 0.31 \\
& Checklist & \textbf{0.90} & \textbf{0.85} & \textbf{0.85} \\
& GYC & 0.56 & 0.77 &  0.61 \\
\midrule
\multirow{4}{*}{\begin{tabular}[c]{@{}l@{}}  Tree-Edit \\ Distance \\ ($\downarrow$ better) \end{tabular}} 
& Masked-LM & 0.93 & 1.67 & 1.40  \\
& PPLM-BoW & 3.76 & 3.35 & 4.00 \\
& Checklist & \textbf{0.38} & \textbf{0.97} & \textbf{0.72} \\
& GYC & 2.87 & 1.79 & 2.69 \\

\bottomrule
\end{tabular}
\caption{Evaluation of Content Preservation (semantic similarity between input sentence and generated counterfactual text) and Tree-Edit Distance (syntactic structure of input sentence and generated counterfactual text) of baselines.}
\label{tab:joint_comparison_aux}
\end{table}
% Small tables : Table naveen-inspired end 

% Page 5, 6, 7

\subsection{Baselines}
We employ baselines with different categories like- dictionary-free, Sentence level counterfactual text, diversity, Enforce \texttt{condition} as shown in Table \ref{tab:baselines}.
We describe the implemented baselines as follows: 
\begin{enumerate}[label=\textbf{(\alph*)}, wide = 0pt, labelindent=0pt]
    \item \textbf{CheckList}: Checklist performs rule-based, lexicon-based and LM based perturbations. It defines a set of pre-defined templates to construct test-cases. This is the only existing model that makes use of LM based perturbations for test-case generation.
    %The LM based peturbations makes use of a [MASK] introduced in \cite{liu2019roberta}. 
    The generated test cases are based on token-substitution. Hence, these provide very rigid perturbations. According to a chosen template, it perturbs the specific part of the template. e.g. name, adjective, or location is taken from a dictionary. In LM-based perturbations, a mask is specified at a certain position in the text and it predicts possible test cases via RoBERTa \cite{liu2019roberta}. For more details, see Appendix B in supplementary document.
    
    %In implementation, we choose a random location for the [MASK] and then generate five counterfactual text samples. Moreover, to incorporate meaningful counterfactual text we choose the mask location after 3 words in any given sentence. We denote this technique as \textbf{CheckList}.    
    
    \item \textbf{BERT}: BERT is a masked language model, where we randomly mask tokens from input sequence and the model learns to predict the correct token. Therefore to compare our approach with LM-based perturbations we considered this as one of the baselines. We took a pre-trained BERT model for our experiments. To generate counterfactual text, we randomly mask one word from the input sequence after skipping the first 2 words as context. We denote this model as \textit{Masked-LM}.
    \item \textbf{PPLM-BoW}: PPLM \cite{dathathri2019plug} uses a bag-of-words to control the text generation. We feed first 2 words of the input sentence as context and the rest of the text is generated. We hypothesize that PPLM-BoW should be unable to generate sentences with similar meaning. This is because it should try to exactly use the words provided in the input text but in wrong order or sense. Due to the interplay between the KL term and the dictionary term it should exhibit low content preservation. 
    \item \textbf{Adversarial Attack}: The work by \cite{michel2019evaluation} formalizes the method of generating meaning-preserving perturbations in the form of adversarial attacks that tend to change the label of the classifier for a given input text. In our setup, we use the implementation of \cite{michel2019evaluation} for adversarial perturbations by Allen AI \cite{Gardner2017AllenNLP}. We denote this model as \textbf{Adversarial Attack}. We use this model for qualitative analysis in the human evaluation section as we intend to compare the plausibility of generated counterfactual text.
\end{enumerate}

\subsection{Datasets}
We perform the experiments on DBpedia \cite{dbpedia}, AgNews \cite{zhang2015character} and Yelp \cite{shen2017style}. All three datasets come from different domains. The details and properties of each dataset taken are described as follows:\\

\noindent
\textbf{DBPedia.} This dataset focuses on locations and organizations. We use publicly available DBpedia dataset\footnote[1]{http://goo.gl/JyCnZq} by \cite{zhang2015character} containing 14 classes. The dataset contains 560K training samples and 70K test samples.

\noindent
\textbf{AgNews.} This dataset focuses on real-world data containing four classes - world, business, sports, and sci-fi categories. Each class contains 30K training samples and 1.9K testing samples. The total number of training samples is 120K and testing 7.6K.

\noindent
\textbf{Yelp.} This dataset focuses on informal text containing reviews. The original YELP Polarity dataset has been filtered in \cite{shen2017style} by eliminating those sentences that exceed 15 words.  We use the filtered dataset containing 250K negative sentences, and 350K positive ones. 

% \textbf{Quora Question Pair.} This dataset focuses on questions.
\subsection{Metrics}
\subsubsection{Generation Quality}
We define various quantitative metrics to measure different aspects of counterfactual text generations. The choice of these metrics is to ensure both the viability of the generated texts as counterfactual text (metric 1,2), and also to examine their plausibility in semantic and syntactic structure (metrics 3,4).
To evaluate the \textit{$K$} counterfactual samples generated from input text $\bx$, the metrics for measuring the proximity of generated counterfactual text samples, diversity, and syntactic structure are as follows: 
% Grammar 

\begin{enumerate}[label=\textbf{(\alph*)}, wide = 0pt, labelindent=0pt]
    \item \textbf{Label-Flip Score}: Label-Flip Score, LFS, is an accuracy-based metric which is used to assess if the generated counterfactual text belong to a different class than the source text class. In our setting, we train an XL-Net~\cite{yang2019xlnet} on different datasets and then compute the labels of the counterfactual text and obtain the Label-flip score. For an input text $\bx_i$ with label $C_i$, we generate $K$ counterfactual text samples and define LFS for this example as:
        \begin{align*}
            \operatorname{LFS}_i = \frac{1}{K} \sum_{k=1}^{K} I(C_i \neq \hat{C_i}) \times 100
        \end{align*}
    %Final $\operatorname{LFS}$ is the mean of these scores across all test examples.
    
    \item \textbf{Diversity Score}: Similar to the Self-BLEU metric proposed by \citet{zhu2018texygen}, we propose a Self-BERT metric using \citet{bert-score} to measure the diversity of generated counterfactual text samples. In our setting, we compute the Self-BERT score between the generated counterfactual text samples. We term this computed Self-BERT score as Diversity Score. A higher value represents lower diversity and vice-versa.
    \item \textbf{Content Preservation}: Content preservation is used to assess the similarity between generated counterfactual with the input sentence. We compute the similarity between input and generated CF on a sentence level using a pre-trained ML model \cite{reimers-2019-sentence-bert}. Since in our setting, we generate $K$ counterfactual text samples, to compute the content similarity, we take a mean of scores over $K$ samples. 
    \item \textbf{Tree-Edit Distance}: To measure the syntactic closeness of generated counterfactual text with the original text, we compare their structures using tree-edit distance \cite{zhangTED}. Tree-edit distance is defined as the minimum number of transformations required to turn the constituency parse tree of a counterfactual generation to that of the original text. This metric is required to ensure that while changing the classifier label, we do not deviate too much from the input text structure. 
\end{enumerate}

\subsubsection{Human Evaluation}
We define qualitative metrics to evaluate the generation quality of the generated counterfactual text with the Adversarial Attack model. We randomly sample outputs from our model and from the baseline model (totaling to 600 samples). For this purpose, 30 annotators rated the generated counterfactual text. The annotators were kept blind to the model that generated the counterfactual text. We adopt five criteria ranging from 1 to 5 (very poor to excellent): \textit{grammatical correctness, plausibility, fluency, sentiment change, content preservation}. Figure \ref{fig:human_eval} shows the human evaluation results. Among GYC and adversarial baseline, GYC achieves the best results on all metrics. In addition to the four properties of counterfactual text, human evaluation clarifies that GYC is able to maintain high grammatical correctness, plausibility, fluency by changing the sentiment and thereby maintaining the balance in generation.

%\textbf{Test-Case Quality.}
% Transfer Strength
% Perc. of Mis\-classified Samples

%\textbf{Model Debiasing.} 
% Flip percentage

\subsection{Results and Discussion}

% Small tables : Table naveen-inspired table 
% \begin{table}[ht]
% \centering
% \scriptsize
% \begin{tabular}{lrccc}
% \toprule
% \multicolumn{1}{l}{Metric} & Model & Dbpedia &  Agnews & Yelp\\
% \midrule
% \multirow{4}{*}{\begin{tabular}[c]{@{}l@{}}  Content \\ Preserv-\\ation \\ ($\uparrow$ better) \end{tabular}} 
% & Masked-LM & 0.89 & 0.84 & 0.83 \\ 
% & PPLM-BoW & 0.18 & 0.20 & 0.31 \\
% & Checklist & \textbf{0.90} & \textbf{0.85} & \textbf{0.85} 
% \bottomrule
% \end{tabular}
% \caption{Evaluation of Content Preservation (semantic similarity}
% \label{tab:joint_comparison_aux}
% \end{table}
% Small tables : Table naveen-inspired end 

% \begin{figure}[ht]
%     \centering
%     \includegraphics[width=2cm,height=1.5cm]{LaTeX/avg_score.png}
%     \caption{Comparison of human evaluation of generated text from Adversarial Attack and GYC.}
%     \label{fig:human_eval}
% \end{figure}

In Table \ref{tab:joint_comparison_main}, we show the results of GYC and other baselines on three different datasets. We observe that GYC produces counterfactuals with close to the highest Label-Flip percentage across all datasets, and establishes high diversity while maintaining content preservation and syntactic structure as compared to baselines like Checklist, Masked-LM, and PPLM-BoW which rely on token-based substitution criteria. \\
% Where as, in comparison the rule-based baselines like Checklist perform poorly on Label-Flip and diversity as they tend to change just on token based on a substitution criteria.

\noindent
\textbf{Performance in flipping the label.}  In Table \ref{tab:joint_comparison_main}, we observe that our label flip score is the highest in Yelp and is significantly high in Dbpedia and AgNews. Checklist and Masked-LM perform poorer than ours. This is because they do not have an explicit learning signal to do the generation that guides them to flip the label. Our model is equipped with differentiable label loss which can guide these generations to flip label. Adversarial Attack can flip the label every time, but the generations are grammatically incorrect and not plausible. Hence, they do not fit our task setting. In PPLM-BoW, the flip score is very low in Yelp due to the reason that it does not have an explicit learning signal to flip the label. In case of AgNews and Dbpedia, the flip score is very high for PPLM-BoW as the number of classes is high in both these datasets and often intermixed like ``Natural Place", ``Village" are two separate classes in Dbpedia. Hence it is justified to show high flip rate in these datasets.\\

\noindent
\textbf{How diverse are the generated samples.} In this metric, we expect that our model provides more diverse samples while maintaining the content of the original text and flipping its label as compared to other models. In Table \ref{tab:joint_comparison_main}, we observe that GYC outperforms all models on Dbpedia and YELP. In AgNews, our diversity is slightly lower than PPLM-BoW. This is expected because PPLM-BoW generates overly diverse sentences that are often unrelated to the original text. \\

\noindent
\textbf{Preserving Content of Input Text.} While we want to generate diverse samples, we also want to make sure that the generations are close to the input text. We expect that Checklist and Masked LM which perform single token-substitution will have high content preservation. Hence, the focus of this metric is not on comparison with these but with PPLM-BoW. In Table~\ref{tab:joint_comparison_aux}, we see that the content preservation of PPLM-BoW is significantly lower than GYC. In PPLM-BoW, most of the sentences are overly diverse and unrelated to the original text.  \\

\noindent
\textbf{Preserving Syntactic Structure.} With this metric, we aim to test that our model does not lose the syntactic structure of the input sentence while flipping the label. In Checklist and Masked-LM the syntactic structure score will be very high since they perform the single token-based substitution. Hence the focus of this metric is not to compare with Checklist and Masked-LM but to compare against PPLM-BoW which also generates free-form text like GYC. In results, we observe that we preserve syntactic structure better than PPLM-BoW and in AgNews our structure preservation is close to that of Checklist.  

We show aggregated results of GYC across all metrics and show that GYC outperforms all the models (see results in Appendix C in supplementary material).

\section{Conclusion}
In this paper, we propose a framework \textit{GYC}, leveraging GPT-2.  We provide the first method that generates test-cases without using any rule-based cues. Experimental results on three datasets show that GYC outperforms existing models and generates counterfactual text samples with a high label-flip score which are both plausible and diverse by preserving content and syntactic structure. 

A next step would be to extend the counterfactual generation for longer sentences, which is currently expensive to run via reconstruction. This will significantly improve the automatic counterfactual generation and thereby robustifying the AI systems.

\bibliography{aaai}

\clearpage

%\input{supplemental}s

% \appendix{Appendix}

% Supplementary Material for: \\ Generate Your Counterfactuals: Towards Controlled\\Counterfactual Generation for Text

% \title{Supplementary Material for: \\ Generate Your Counterfactuals: Towards Controlled\\Counterfactual Generation for Text}
% \maketitle

\begin{appendices}
\section{Appendix A: Model Details}
Our model consists of multiple losses to generate counterfactuals with four properties. In this section we will describe the implementation details with hyperparameters required for different losses.

\subsection{Reconstruction Loss Implementation}
In reconstruction loss, we tried to regenerate the input sentence such that we can perturb the hidden state of input for generation of counterfactuals. While this is straightforward, while using GPT-2 the inference is a hard task. 

We explain the ways we do inference on GPT-2. \\

\noindent
\textbf{Method 1: Naive Training}
In naive training, we try to maximize the presence of every given $\displaystyle x_{i}$ in the sentence all at once right from the beginning of the training. \\
\begin{equation*}
    \displaystyle \text{Loss}\ =\ \sum ^{T}_{t=1}\log \ p\left( x_{t} |\tilde{H}_{< t}\right) \ 
\end{equation*}

The main problem with this approach has been that the gradient is maximum at the last word/ hidden state. Hence it gets stuck in local minima. Therefore the remaining sentence can not be reconstructed due to weak gradients. So we could not adopt this for reconstruction loss.\\

\noindent
\textbf{Method 2: Curriculum Training}
In curriculum training, we try to maximize the presence of \ given $\displaystyle x_{i}$ in the sentence in left to right order. So we first try to reconstruct $\displaystyle x_{1\ }$then $\displaystyle x_{2}$ and so on. 
\begin{equation*}
\displaystyle \text{Loss}\ =\ \sum ^{\ K}_{t=1}\log \ p\left( x_{t} |\tilde{H}_{< t}\right) \ 
\end{equation*} where $\displaystyle K$ is chosen according to the number of iterations.

The main problem with this approach has been the length of the each phase is hard-coded which makes it very slow and empirically performs poorly.\\

\noindent
\textbf{Method 3: Adaptive Curriculum Training (This is the working solution)}
\begin{equation*}
    \displaystyle \text{Loss}\ =\ \sum ^{\ K}_{t=1}\log \ p\left( x_{t} |\tilde{H}_{< t}\right) \
\end{equation*} where $\displaystyle K$ is chosen on the basis of how many words are already reconstructed

In adaptive curriculum training, we try to maximize the presence of \ given $\displaystyle x_{i}$ in the sentence in left to right order. So we first try to reconstruct $\displaystyle x_{1\ }$then $\displaystyle x_{2}$ and so on. 

We go to the next phase of curriculum as soon as previous word is regenerated. If previous word goes out of the construction, then we roll-out of the curriculum. This helps in doing the reconstruction of the input sentence in a reliable way. Hence we follow this approach in the paper. 

While reconstruction is of utmost importance when we perturb to generate $\bx$, we control the strength of reconstruction or proximity loss by assigning a proximity\_loss\_weight. This weight is set to 1 during initial phases of training. After the sentence has been reconstructed, this weight is set to 0.02 so that perturbation can effective and not regenerate the original sentence.  

\begin{table*}
% \centering
\small
\begin{tabular}{@{}|p{0.17\textwidth}|p{0.17\textwidth}|p{0.17\textwidth}|p{0.17\textwidth}|p{0.17\textwidth}|@{}}

\hline 
& & & & \\
\multicolumn{1}{|c|}{\textbf{Input Sentence}} & \multicolumn{1}{|c|}{\textbf{GYC}} & \multicolumn{1}{|c|}{\textbf{Checklist}} & \multicolumn{1}{|c|}{\textbf{Masked-LM}} & \multicolumn{1}{|c|}{\textbf{PPLM-BoW}} \\

\hline 
& & & & \\
I would recommend this place again & I would \textbf{not} recommend this place &  I would recommend this \textbf{site} again &  I would recommend your \textbf{place} again & I would \textbf{like to thank} this \textbf{user} \\
& & & & \\

\hline 
& & & & \\
The wait staff is lazy and never asks if we need anything & The wait staff and \textbf{volunteers are always ready to offer their assistance} & The wait \textbf{list} is lazy and never asks if we need anything & The wait staff is lazy and never asks if we need \textbf{?} & The wait \textbf{is over - it's a real thing. You will receive}\\
& & & & \\

\hline 
& & & & \\
1st time burnt pizza was horrible & 1st time burnt pizza \textbf{is delicious} & 1st time burnt pizza was \textbf{eaten} & 1st time burnt pizza was \textbf{sold} & 1st time, \textbf{this is a great place}\\
& & & & \\

\hline 
& & & & \\
It is extremely inconvenient & It is exceedingly \textbf{easy} & It is \textbf{not} inconvenient & It is extremely \textbf{beautiful} & It is \textbf{very hard to tell}\\
& & & & \\

\hline 
& & & & \\
The beer is awful & The beer is \textbf{amazing} & The beer is \textbf{delicious} & The beer \textbf{smelled} awful & The beer \textbf{is a blend of}\\
& & & & \\

\hline 
& & & & \\
The return policy here is ridiculous & The return policy here is \textbf{simple} & The return \textbf{fee} here is ridiculous & The return policy \textbf{which} is ridiculous & The return \textbf{of a legendary, but rarely}\\
& & & & \\

\hline 
& & & & \\
I have a great experience here & I have this \textbf{dish} & I \textbf{got} a great experience here & I have \textbf{one} great experience here & I have \textbf{the following issue with the latest}\\
& & & & \\

\hline 
& & & & \\
You should give this place zero stars & You should \textbf{be able to} give stars & You should give this \textbf{game} zero stars & You should give \textbf{the} place zero stars & You should \textbf{read it. This is an extremely}\\
& & & & \\

\hline 
& & & & \\
Our waitress was very friendly & Our waitress was \textbf{not amused} & Our \textbf{waiter} was very friendly & Our waitress was very \textbf{!} & Our waitress was \textbf{super friendly and friendly}\\
& & & & \\

\hline 
& & & & \\
I had a great experience here & I had \textbf{to write about this} & I had a great experience \textbf{there} & I had a great \textbf{time} here & I had a \textbf{great time on this one}\\
& & & & \\

\hline
\end{tabular}
\caption{Illustration of samples generated by GYC in comparison with Checklist, Masked-LM and PPLM-BoW}
\label{tab:example_perts_new}
\end{table*}

\subsection{Differentiable Loss Implementation}
For the differentiable loss, we train a sentiment model on different datasets used in our experiments. The input to sentiment model is logits ($p_{v}$) of the generated text after reconstruction step ($\tilde{\bH}$). In order to have this work with GPT-2, the vocabulary of sentiment model and GPT-2 has to be consistent. 

For this purpose, we choose our sentiment model to be composed of an EmbeddingBag layer followed by a linear layer. We take the mean value of the bag-of-embeddings ($e_{v}$) over the whole vocabulary, $v$. We represent the logits of classifier as follows -  

\begin{equation*}
    \text{logits}\ = \operatorname{MLP}\left(\sum _{t}\sum _{v} p^{t}_{v} e_{v}\right)
\end{equation*}

\begin{figure}[!ht]
    \centering
    \includegraphics[width=1.0\linewidth,height=4.5cm]{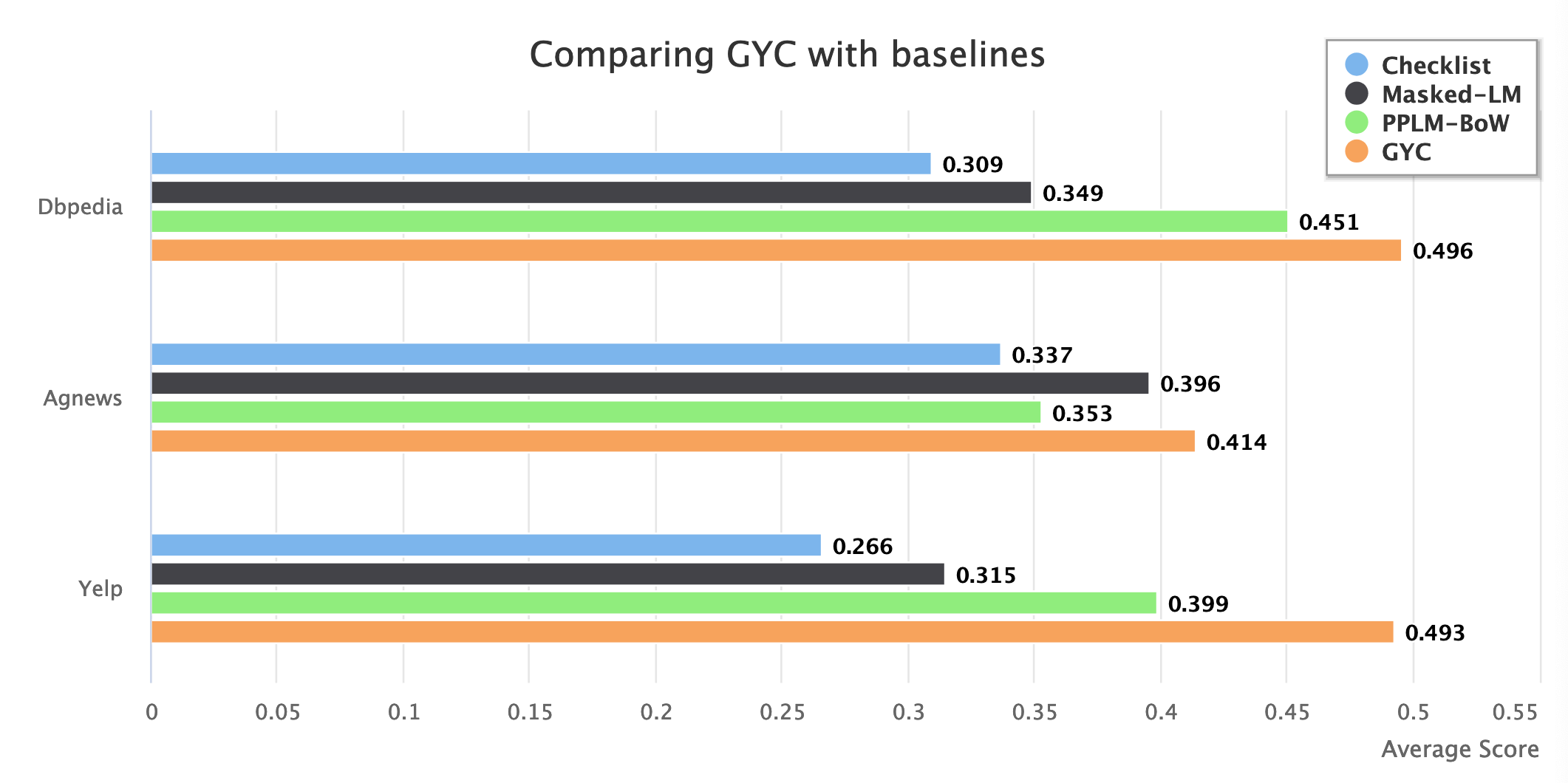}
    \caption{Comparison of GYC with existing approaches on average score of all four metric.}
    \label{fig:human_eval_new}
\end{figure}

\subsection{Reward Loss Implementation}
For thee reward loss implementation, we generate reward for non-differentiable score function. The implementation is done for capturing reward based-on NER, SRL and Sentiment Label-based. 

\begin{itemize}
\item We compute reward for each sentence $\displaystyle x^{k}$.
\begin{itemize}
\item NER-based reward\textbf{ }is\textbf{ }computed as $\displaystyle p\left(\text{NER-tag} \ in\ x^{k}\right)$
\item SRL-based reward\textbf{ }is\textbf{ }computed as $\displaystyle p\left(\text{SRL-tag} \ in\ x^{k}\right)$
\item Label-based reward is computed as \ $\displaystyle p\left(\text{Label-tag} \ in\ x^{k}\right)$
\end{itemize}
\end{itemize}

\subsection{Entropy}

To generate samples with enough diversity, we employ entropy. To control entropy, we use an entropy\_weight parameter. In our experiments entropy is set to 0.08.  
\section{Appendix B: Baselines Implementation}

\textbf{Checklist}
In Checklist, we use the LM based perturbations employing RoBERTa model. We generate a word by choosing [MASK] at a random location in the input text. This mask is chosen after ignoring the first three tokens of the input sentence, to make sure that we mask a relevant part of the sentence for replacement.

\section{Appendix C: Results}

\subsection{Quantitative Results}
We aggregate the results of GYC averaged over all four metrics to assess the performance with baselines in Figure \ref{fig:human_eval_new}. We observe that, GYC has the highest average score compared to the baselines on all datasets. 

%\newpage

\subsection{Qualitative Results}
We show generated counterfactual samples from GYC and the baselines for comparison in Table \ref{tab:example_perts_new}. 

\end{appendices}

\end{document}